\documentclass[runningheads]{llncs}
\usepackage[T1]{fontenc}
\usepackage{graphicx}
\usepackage{version}
\usepackage{multirow}
\usepackage{rotating}
\usepackage{amssymb,amsmath}
\usepackage{mathtools}
\DeclarePairedDelimiter\floor{\lfloor}{\rfloor}
\usepackage{hyperref}
\usepackage{enumerate}
\usepackage{xcolor}
\newcommand{\fig}{Fig.~}
\newcommand{\SSS}{\mathcal{S}}
\newcommand{\GG}{\mathcal{G}}
\newcommand{\omni}{360°~}
\newcommand{\llbracket}{[\![}
\newcommand{\rrbracket}{]\!]}

\usepackage{color}

\begin{document}
\title{Deep Spherical Superpixels}
\author{R{\'e}mi Giraud\inst{1} %\orcidID{0000-1111-2222-3333} 
\and
Micha{\"e}l Cl{\'e}ment\inst{2} %\orcidID{1111-2222-3333-4444} 
}
\authorrunning{R. Giraud et al.}
\institute{
Univ. Bordeaux, Bordeaux INP, IMS, CNRS UMR 5218, France.\\ 
\email{remi.giraud@ims-bordeaux.fr} 
\and
Univ. Bordeaux, Bordeaux INP, LaBRI, CNRS UMR 5800, France.
\email{michael.clement@labri.fr}
}
\maketitle              % typeset the header of the contribution
\begin{abstract}
Over the years, the use of superpixel segmentation has become very popular in various applications, serving as a preprocessing step to reduce data size by adapting to the content of the image, regardless of its semantic content.
While the superpixel segmentation of standard planar images, captured with a 90° field of view, has been extensively studied, 
there has been limited focus on dedicated methods to omnidirectional or spherical images, captured with a 360° field of view.
In this study, we introduce the first deep learning-based superpixel segmentation approach tailored for omnidirectional images called DSS
(for Deep Spherical Superpixels).
Our methodology leverages on spherical CNN architectures and the differentiable $K$-means clustering paradigm for superpixels, 
to generate superpixels that follow the spherical geometry.
Additionally, we propose to use data augmentation techniques specifically designed for 360° images, enabling our model to efficiently learn from a limited set of annotated omnidirectional data.
Our extensive validation across two datasets demonstrates that taking into account the inherent circular geometry of such images into our framework improves the segmentation performance over traditional and deep learning-based superpixel methods.
Our code is available online\footnote{\url{https://github.com/rgiraud/dss}}.
\end{abstract}

\section{Introduction}

The vast majority of computer vision methods are tailored
for standard RGB images, \emph{i.e.}, 
captured with a standard 90° field of view (FoV).
However, acquisition devices with wider FoV
have become more and more popular in the recent years.
In particular, omnidirectional images with a 360°$\times$180° FoV
are very interesting to capture the entire environment of a scene.
Over the literature, such imagery may be equally referred 
as omnidirectional, spherical, 360°, or even panoramic.
Naturally, such acquisition introduces distortions when projecting the capture 
on a planar 2D image. %\cite{zorin1995correction}.
Nevertheless, many dedicated methods have been successfully applied on these images, for example for scene reconstruction~\cite{sun2019horizonnet},
semantic segmentation~\cite{yang2020omnisupervised} for autonomous driving,
or in the context of mixed or virtual reality~\cite{da2019dense}.

To efficiently apply deep learning-based architectures to these images,
a few adjustments must be made to consider their specific geometry.
For instance, the input images are horizontally circular so the pixels of the first column should be considered spatially adjacent to the pixels of the last column.
Some methods explicitly take into account these geometrical properties, for instance with spherical convolutional neural networks (SCNNs) that have demonstrated higher performance on \omni images than standard CNNs~\cite{cohen2018spherical}.
Nevertheless, as for any deep learning-based method,
a significant amount of annotated data is necessary for an efficient training, especially when tackling segmentation applications.

For regular standard images, various segmentation datasets are available 
with different content, resolution or precision in the annotations. 
However, only a few spherical image datasets are available, 
such as SUN360~\cite{xiao2012recognizing} 
or Matterport3D~\cite{chang2017matterport}.
Moreover,
due to the tediousness of a pixel-wise semantic segmentation process, 
they generally only provide layout, depth or camera pose information~\cite{da20223d}.
In the context of autonomous driving, 
many datasets contain pixel-wise semantic annotations 
but the FoV is generally limited to standard rectangular acquisition~\cite{Geiger2012CVPR,cordts2016cityscapes},
or the images are captured by a fisheye lense introducing other distortions~\cite{yogamani2019woodscape}. %woodscape
Hence, deep learning segmentation methods that are applied 
to \omni imagery may highly necessitate specific data augmentation strategies \cite{sun2019horizonnet,yang2020omnisupervised}.

In a more general context of image segmentation methods, 
non-semantic decompositions into superpixels offer numerous benefits.
These methods regularly group pixels into homogeneous and connected regions, respecting the image contours. 
They have mainly been popularized by SLIC~\cite{achanta2012}, 
a simple method that uses a locally constrained iterative $K$-means clustering, computed on color and position features. 
Then, many derived methods have been proposed,
such as the non-iterative SNIC method~\cite{achanta2017superpixels}, 
LSC~\cite{chen2017} which expands the feature space of SLIC, 
or SCALP~\cite{giraud2018_scalp} that computes a color consistency along the path between a pixel and the centroid of its superpixel. 
Other methods like GMMSP~\cite{Ban18} propose different strategies, such as using a Gaussian Mixture Model.

The first superpixel method tailored for spherical images
was proposed in~\cite{wan2018}, extending SLIC.
The spherical geometry is considered in 
the clustering distance, that is computed using the 3D positions
of pixels on the sphere.
The produced superpixels are regular on the 3D sphere domain and are able adapt to the
distortions of objects induced by the projection on the 2D planar image, 
leading to higher segmentation performance compared to planar methods.
Following, many planar superpixel algorithms have had their omnidirectional counterparts, 
such as SSNIC~\cite{da2021fast}, SphLSC and SphSPS (or SphSCALP)~\cite{giraud2023sphsps}.

Nevertheless, over the years, all these traditional approaches have started
to report saturated performance over the segmentation benchmarks.
With the Superpixel Sampling Network (SSN) method \cite{jampani2018superpixel}, 
a first deep learning framework has been proposed to compute a segmentation into superpixels.
SSN and following methods, \emph{e.g.},~\cite{yang2020superpixel}, enable to improve the segmentation accuracy by computing more advanced features,
with the use of a CNN trained on higher-level annotated segmentations (for example from semantic segmentations).
However, these deep learning methods have only been designed 
for standard planar images.

\subsubsection{Contributions}

In this work, we propose the first deep learning-based method
called Deep Spherical Superpixels (DSS),
able to segment omnidirectional images into spherical superpixels.
The contributions of this work are listed as follows:
\begin{enumerate}[i]
\item  We introduce the first deep learning-based superpixel segmentation method tailored for omnidirectional images, leveraging spherical CNN architectures and the differentiable $K$-means superpixel algorithm ;
\item  We make use of specific data augmentation strategies designed for \omni %spherical %omnidirectional 
images, whose effectiveness is demonstrated through an ablation study ;
\item We comprehensively evaluate the proposed method against state-of-the-art approaches, including both traditional planar and spherical approaches as well as deep learning-based methods, evaluated for the first time on the spherical superpixel segmentation task ;
\item We propose a quantitative validation on the Panorama Segmentation Dataset (PSD)~\cite{wan2018}, the reference for spherical superpixels,
on initial and noisy images,
and also on a newly considered omnidirectional road dataset, Wild PAnoramic Semantic Segmentation (WildPASS)~\cite{yang2021capturing} ;
\item The source code of our method is made available to the research community.
\end{enumerate}

%\section{Deep \omni superpixels}
\section{Deep Spherical Superpixels Method}

In this Section, we introduce our proposed Deep Spherical Superpixels (DSS) method.
First, we present the Superpixel Sampling Network (SNN)~\cite{jampani2018superpixel} framework that we use as basis for our method (Section~\ref{subsec:ssn}).
Then, we detail the
\omni coordinates system (Section~\ref{subsec:geometry}) 
and our modifications of SSN to generate spherical superpixels (Section~\ref{subsec:360network}).
Finally, we present the 360°-specific data augmentation 
used to enable our model to efficiently learn from a limited set of annotated omnidirectional data (Section~\ref{subsec:DA}).

\subsection{Superpixel Sampling Network}
\label{subsec:ssn}

In the superpixel segmentation literature, the Simple Linear Iterative Clustering (SLIC) algorithm is one the most simple yet accurate method~\cite{achanta2012}.
It performs a locally constrained $K$-means clustering starting from a regular sampling grid. 
This clustering relies on a spatial and a color distance between each pixel and a superpixel centroid.
Although SLIC is interesting for its rapidity and ease to use, its clustering accuracy can be limited since it is only based on RGB or Lab image features. 

In \cite{jampani2018superpixel}, an end-to-end framework is proposed using 
a convolutional neural network (CNN) trained to
learn how to provide more advanced features as input 
to a differentiable SLIC clustering algorithm.
The network is trained to produce superpixels that are contained into higher-level annotated segmentations (for example from semantic segmentations).
In particular, the integration of SLIC into a deep learning framework is possible in a differentiable manner by considering \textit{soft} mappings of pixels to superpixels.
At inference time, the final \textit{hard} mapping, associating a pixel to a unique superpixel, is only computed to generate the final segmentation.

The SSN model takes as input images of size $N = h \times w$, represented with 5 channels corresponding to \textit{Lab} color features (3 channels) and $xy$ pixel coordinates (2 channels).
The goal of the model is to learn deep features that are more suitable to perform a differentiable clustering into superpixels.
To achieve this, the SSN model uses a CNN composed of three blocks, each with two convolutional layers, batch normalization and ReLU activation,
with a max pooling layer applied after each block.
For the output, feature maps of each block are upsampled to the original image size (if necessary, for the second and third blocks) and concatenated.
The original \textit{Lab} and \textit{xy} features are also concatenated into the output feature maps, resulting in $D$-dimensional pixel features (\textit{i.e.,} 5 channels from input features and $D-5$ learned deep features).
In practice, the SSN model used $D=20$
in their experiments.
For more details about this architecture, the reader can refer to~\cite{jampani2018superpixel}.

% Loss description.
These learned features are then fed to the aforementioned differentiable clustering to compute soft assignments of pixels to superpixels.
These soft assignments are in turn used to compute a loss function tailored for the desired superpixel properties.
For example, to obtain superpixels matching semantic segmented objects, the loss is comprised of two terms: (i) a pixel-wise cross-entropy term between ground-truth semantic segmentation and predicted soft superpixels and (ii) a compactness term which encourages superpixels to have low spatial variance.

The method is therefore end-to-end trainable and can learn deep pixel features tailored for subsequent superpixels properties.
In the following, we present how to adapt this approach to the specific case of generating spherical superpixels for omnidirectional images.

\subsection{Spherical Geometry\label{subsec:geometry}}

The projection system between the planar equirectangular 2D space and the 3D spherical space is depicted in \fig\ref{fig:3D_coordinates}.
This relationship can be understood through the projection of vertical and horizontal coordinates of the plane onto the sphere's meridians and latitude circles.
This process creates a spherical image where the width $w$ is double the height $h$.
It implies a horizontal continuity in the planar image domain that characterizes omnidirectional images.
Hence, each pixel $p=[j,i]$ in the 2D space matches a 3D point $X=[x,y,z]$ on the unit sphere following the equations:

\begin{figure}[t]
\centering
\includegraphics[width=\textwidth]{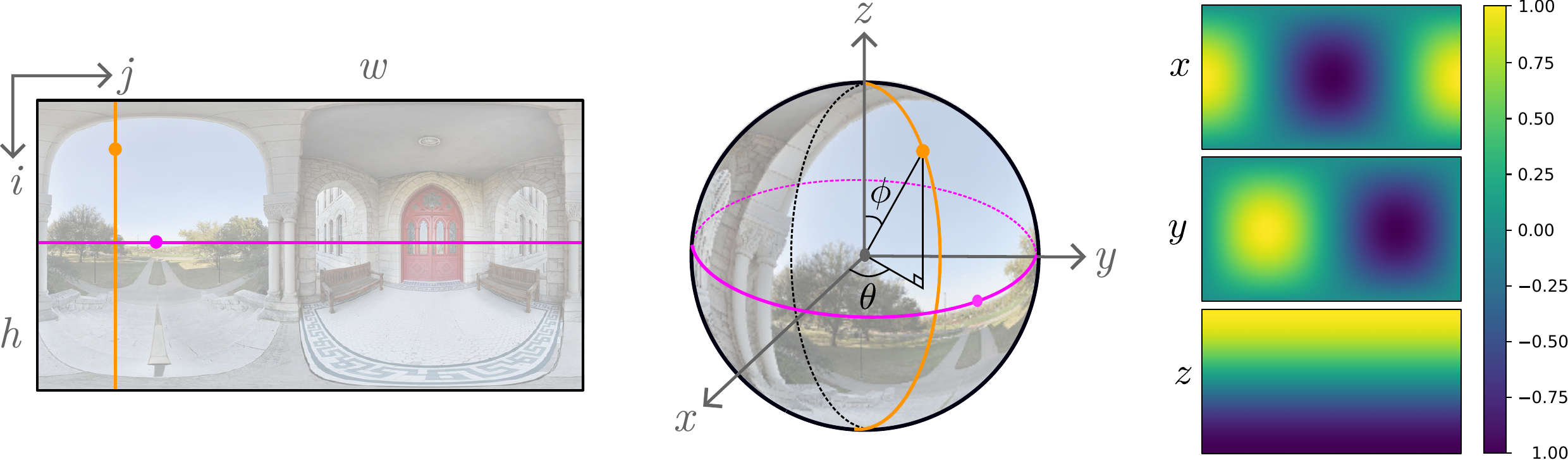} \\[-1.5ex]
\caption{2D Planar and 3D spherical system coordinates. A pixel at position $[j,i]$ in the 2D space is mapped to a 3D point $[x,y,z]$ on the unit sphere following \eqref{xyz}.
This point can also be represented by its respective azimuthal and polar angles $\theta$ and $\phi$.
}
\label{fig:3D_coordinates}
\end{figure} \vspace{-0.1cm}

\begin{equation}
p =
\begin{bmatrix}
j = \floor{\frac{\theta w}{2\pi}} \\[-0.5ex] \\[-0.25ex]
i = \floor{\frac{\phi h}{\pi}}
\end{bmatrix}
\hspace{0.2cm}
\leftrightarrow 
\hspace{0.2cm}
X =
\begin{bmatrix}
x = \text{sin}(\frac{y\pi}{h})\text{cos}(\frac{2x\pi}{w}) \\[0.25ex]
y = \text{sin}(\frac{y\pi}{h})\text{sin}(\frac{2x\pi}{w}) \\[0.25ex]
z = \text{cos}(\frac{y\pi}{h})
\end{bmatrix}
,   \label{xyz}
\end{equation}
where $\theta=\text{arctan2}({y,x})$ is the azimuthal angle, and $\phi=\text{arccos}(z)$ is the polar angle.
Note that this mapping of coordinates considers that $j\in [-\frac{w}{2},\frac{w}{2}]$ so to map $x$ to $[0,w]$, we have
%it is registered %in the image space 
$x \leftarrow x+w$ when $x\leq0$.

\subsection{Spherical Superpixel Clustering Network\label{subsec:360network}}

In this Section, we describe our adaptation of the $K$-means differentiable superpixel clustering network~\cite{jampani2018superpixel} to provide superpixels that are regular over the spherical domain.
We use the same CNN architecture as basis for our method.

\subsubsection{Features and superpixels initialization}

As input for the CNN, we use the $Lab$ color features of the $N=h{\times}w$ pixels, denoted as $F_c\in\mathbb{R}^{N{\times}3}$.
The pixel coordinates are also given as input, but instead of the 2D pixel positions, 
we provide the 3D spherical coordinates $F_s\in[-1, 1]^{N{\times}3}$.
To match the coordinates domain,
we normalize the Lab features $F_c$ to also lie in $[-1, 1]$.

\begin{figure}[t]
\centering
{\scriptsize 
\begin{tabular}{@{\hspace{0mm}}c@{\hspace{1mm}}c@{\hspace{3mm}}c@{\hspace{1mm}}c@{\hspace{0mm}}}
\includegraphics[height=0.16\textwidth]{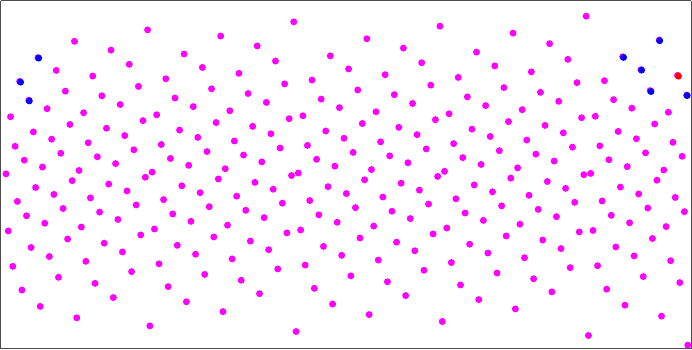}&
\includegraphics[height=0.16\textwidth]{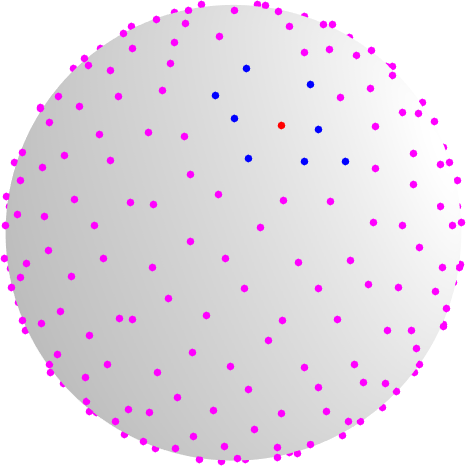}&
\includegraphics[height=0.16\textwidth]{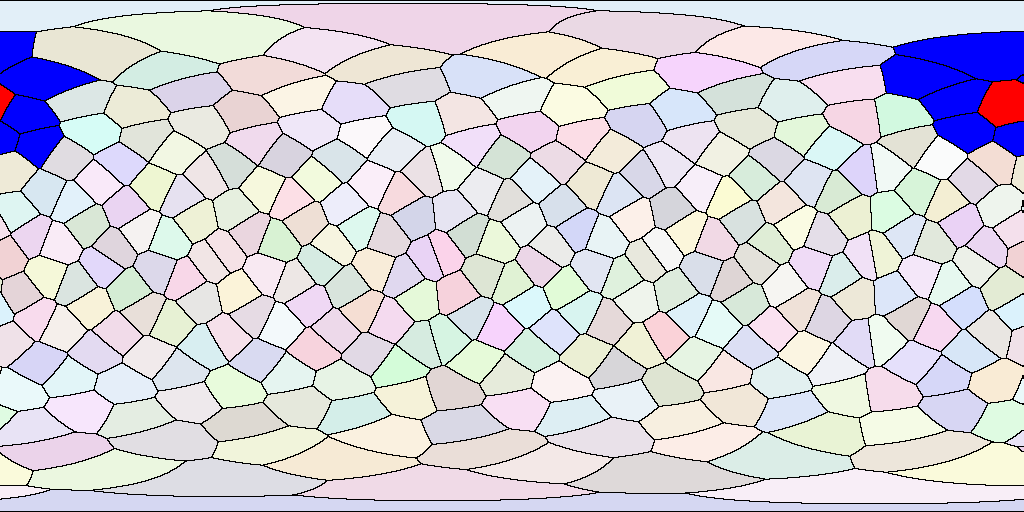}&
\includegraphics[height=0.16\textwidth]{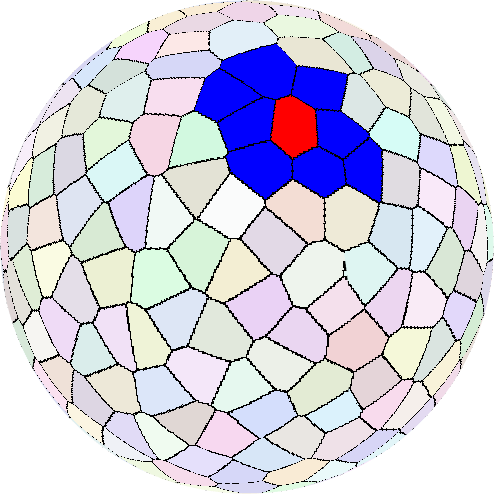}\\[0.5ex]
\multicolumn{2}{c}{(a) Hammersley spherical sampling}&
\multicolumn{2}{c}{(b) Initial label map}\\[-1.5ex]
\end{tabular}
}
\caption{Spherical label map initialization.
(a) A Hammersley sampling with $K=300$ centroids points is computed on the unit sphere. Note the lower sampling density at the vertical borders, corresponding to the sphere's poles. 
(b) Corresponding label map, where each pixel is associated to the closest Hammersley barycenters, producing regular regions on the sphere.
The 8 neighbors of the red superpixel (closest in the spherical space) are represented in blue.} \vspace{-0.15cm}
\label{fig:initialization}
\end{figure}

With classical 2D images, superpixel clusters are usually initialized by a regular sampling on the 2D grid.
However, this strategy is not ideal with omnidirectional images as it does not respect the underlying 3D geometry.
To overcome this issue, many spherical sampling strategies have been compared for superpixel clusters initialization~\cite{da2021fast,giraud2023sphsps}.
In our proposed DSS method, as in~\cite{giraud2023sphsps}, we use Hammersley sampling~\cite{wong1997sampling} to rapidly provide an appropriate set of $K$ 3D points that are uniformly distributed on the unit sphere (see \fig\ref{fig:initialization}(a)).
From this set of 3D points, we define an initial label map 
by a nearest neighbor computation on the 3D pixel position $X$
(see \fig\ref{fig:initialization}(b)).
Such spherically uniform sampling implies a sparser 2D sampling on the planar image near vertical borders.
Classical planar methods that consider an initial regular grid would produce
very irregular oversegmentation around the sphere's poles, as shown later in Section~\ref{subsec:results}.
From this label map, we extract the initial superpixel features with an average pooling.

\subsubsection{Neighborhood-based distance}

In the original SLIC method~\cite{achanta2012}, 
the {$K$-means} clustering is locally constrained so each superpixel can only aggregate a pixel in a fixed sized square window centered on the superpixel barycenter.
For efficient implementation purposes, 
the $K$-means-based differentiable clustering of SSN~\cite{jampani2018superpixel}
slightly differs by iteratively computing the pixel association within the 9-th superpixel neighborhood of the initialization map. 
Therefore, the core of the clustering distance computation is geometry-agnostic, once the superpixel neighbors are identified.
In our context, we can compute for each superpixel a $n$-th neighborhood with a nearest neighbors distance on their 3D barycenters in the spherical space.
Such neighborhood is represented in \fig\ref{fig:initialization}.

Therefore, contrary to the planar square sampling, our method can define without ambiguity a $n\in\llbracket 0,N\rrbracket$-th neighborhood.
In practice, we use a $n=9$ neighborhood, as in SSN.

\subsubsection{Horizontally circular clustering}

\omni images are particularly characterized by their
horizontally circular nature. 
This aspect is not considered in standard CNNs, which typically use zero padding strategies for convolutions and where the final receptive field may be also lower than the image dimension.
In the context of spherical superpixel clustering, without any semantic aggregation of clustered regions,
using standard convolutions is 
highly irrelevant
since we would observe a discontinuity in the segmentation at the image borders.

\begin{figure}[t]
\centering
{\scriptsize 
\begin{tabular}{@{\hspace{0mm}}c@{\hspace{1mm}}c@{\hspace{3mm}}c@{\hspace{1mm}}c@{\hspace{0mm}}}
\includegraphics[height=0.16\textwidth]{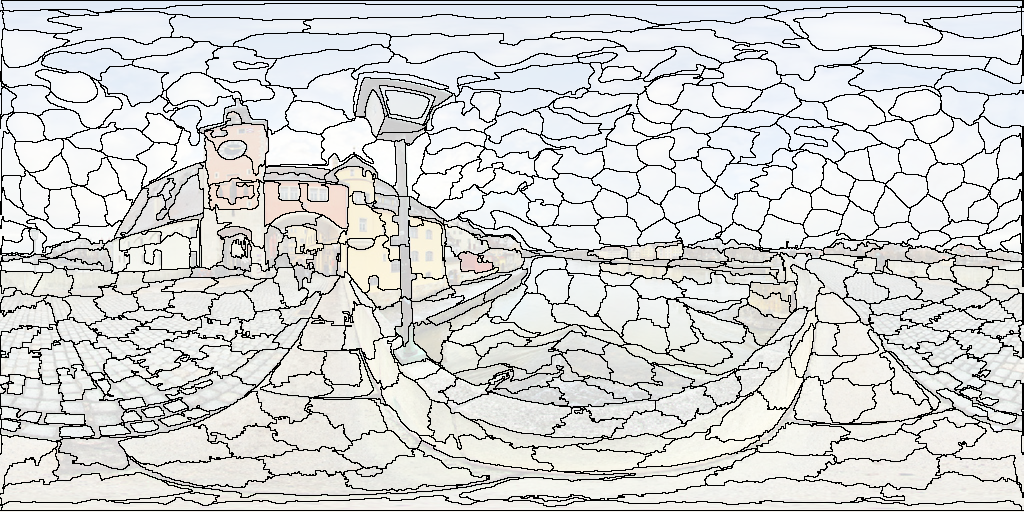}&
\includegraphics[height=0.16\textwidth]{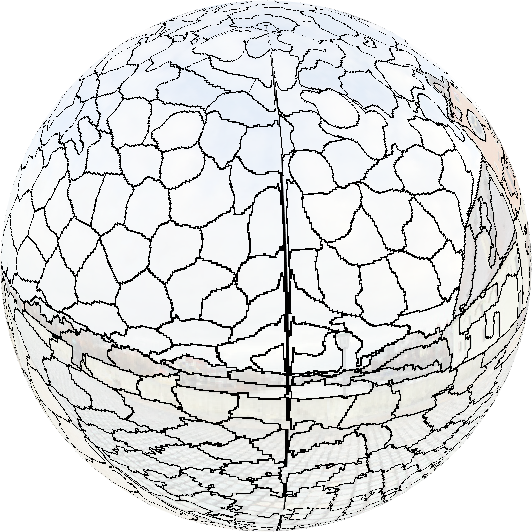}&
\includegraphics[height=0.16\textwidth]{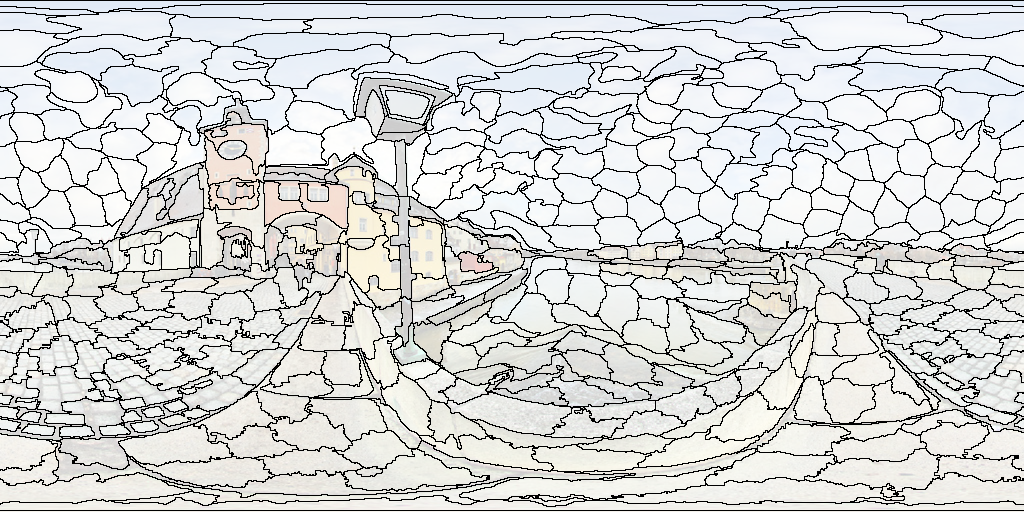}&
\includegraphics[height=0.16\textwidth]{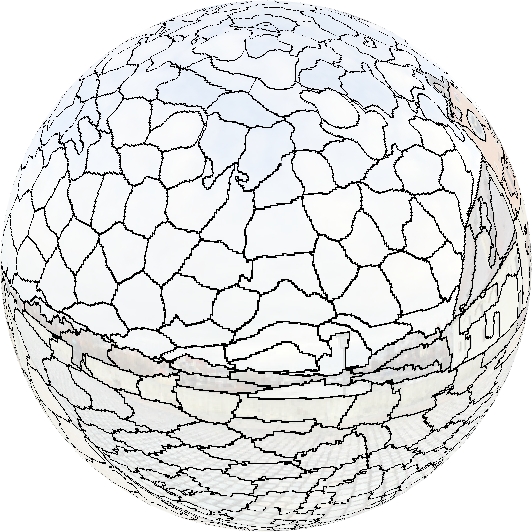}\\[0.5ex]
\multicolumn{2}{c}{(a) With zero padding}&
\multicolumn{2}{c}{(b) With circular padding}\\[-1ex]
\end{tabular}
}
 \caption{Impact of the circular padding on the superpixel segmentation.
(a) With standard zero padding, the CNN features of pixels at $j=0$ and $j=w-1$ 
are not consistent with neighborhood, leading to a vertical border in the spherical space, as for planar methods.
(b) With circular padding, the features remain consistent on the borders and the method is able to fully consider the geometry of the omnidirectional images. 
 }
 \label{fig:circular}
\end{figure}

For example, \fig\ref{fig:circular}(a) shows the result obtained by using a standard zero padding strategy in the CNN layers.
With $3{\times}3$ convolution kernels, the features extracted for pixels at $j=0$ and $j=w-1$ are not consistent with the ones of their neighborhood, which disrupts the selection of their closest superpixel among the $9$ closest.
When computing the hard clustering association, border pixels are generally associated to a disconnected region resulting in the appearance of an artificial vertical border in the spherical space, as for planar methods.
%(see Figure \ref{fig:circular}(a)).

To take into account this horizontally circular geometry into our model, 
we propose to use a \textit{spherical CNN} with a more natural circular padding strategy, as in~\cite{wang2018omnidirectional,schubert2019circular}.
Our spherical CNN uses a horizontal circular (or periodic) padding of half size of the kernel at each step requiring padding (convolutional or max pooling layers).
A replicate padding strategy is used for vertical padding. 
Hence, the spherical CNN is fully able to preserve the \omni geometry in the final clustering and to compute relevant features at the borders.
Note that other strategies may be possible, such as applying a large input circular padding as a preprocessing~\cite{shi2015deeppano}, but with many successive convolutions, this leads to handle significantly larger images, and thus to higher memory and time consumption.

\subsubsection{Loss function} Deep pixel features from our spherical CNN are fed to the differentiable clustering method to produce soft assignments $\SSS_\mathrm{soft}$ of spherical superpixels.
As in SSN, the model is trained with a loss comprised of a pixel-wise cross-entropy with ground-truth segmentation $\GG$ denoted $L_\mathrm{seg}$, and a compactness term $L_\mathrm{compact}$ to enforce superpixels with low spatial variance:
\begin{equation}
    L = L_\mathrm{seg}(\GG,\SSS_\mathrm{soft}) + \lambda L_\mathrm{compact}(F_s,\SSS_\mathrm{soft}) .
    \label{loss}
\end{equation}

\subsubsection{Region connectivity}

After training, to compute the final superpixel segmentation of an image,
a last step ensures the connectivity of the produced regions as for most superpixel clustering methods~\cite{achanta2012,jampani2018superpixel}.
This is simply done by aggregating the smallest disconnected regions to the largest and nearest one but taking into account the circular aspect.

\subsection{360\textsuperscript{o}-Specific Data Augmentation\label{subsec:DA}}

In the context of \omni imagery,
the lack of extensive image datasets with segmentations makes it hard to train neural networks efficiently.
To mitigate this data limitation, the use of data augmentation strategies is crucial. 
While simple augmentation techniques such as flips, blurs,
and noise addition are applicable, they may be insufficient to 
provide enough diversity to the training process.
However, many other conventional data augmentation strategies may alter the intrinsic \omni geometry and should not be used for such images.
For instance, rotations or crops, as used in SSN~\cite{jampani2018superpixel}, compromise the spherical geometry, 
leading to the loss of the horizontal mirror effect and the spatial distortion of the 2D label map.
Using such augmentation techniques would lead the model to learn to provide irregular superpixels in the spherical space with artificial vertical borders at the edges, as for planar methods (see \fig\ref{fig:circular}(a)).

\begin{figure}[t]
{\scriptsize
\begin{tabular}{@{\hspace{0mm}}c@{\hspace{1mm}}c@{\hspace{1mm}}ccc@{\hspace{0mm}}}
%\textbf{Standard data augmentation} & 
 & &\multicolumn{1}{c}{\small Blur \& noise} &
 \multicolumn{1}{c}{\small Crop \& mirror}&
 \multicolumn{1}{c}{\small Panoramic stretch}\\[0.5ex]
%\rotatebox{90}{\hspace{0.3cm} Initial image}
&\rotatebox{90}{\hspace{0.1cm}Initial image} &
\includegraphics[width=0.3\textwidth]{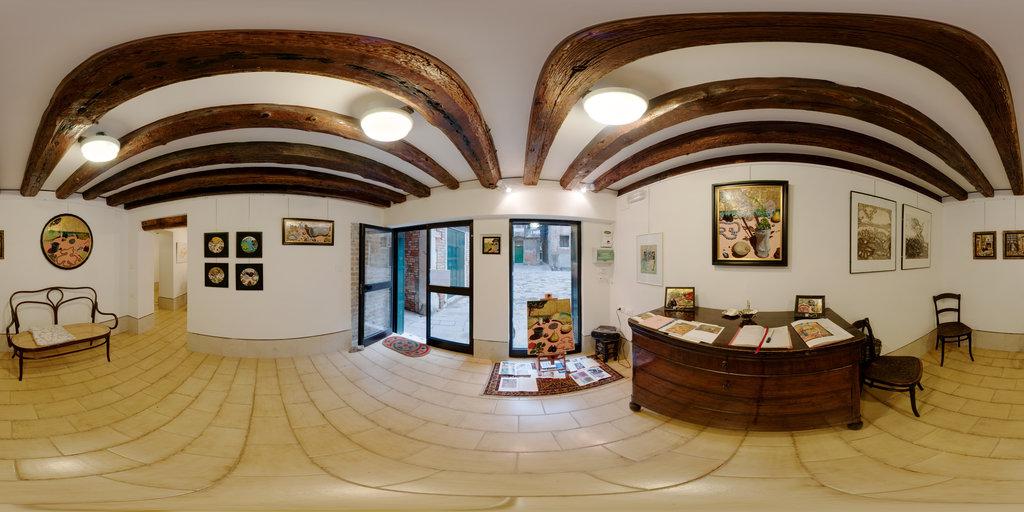}&
\includegraphics[width=0.3\textwidth]{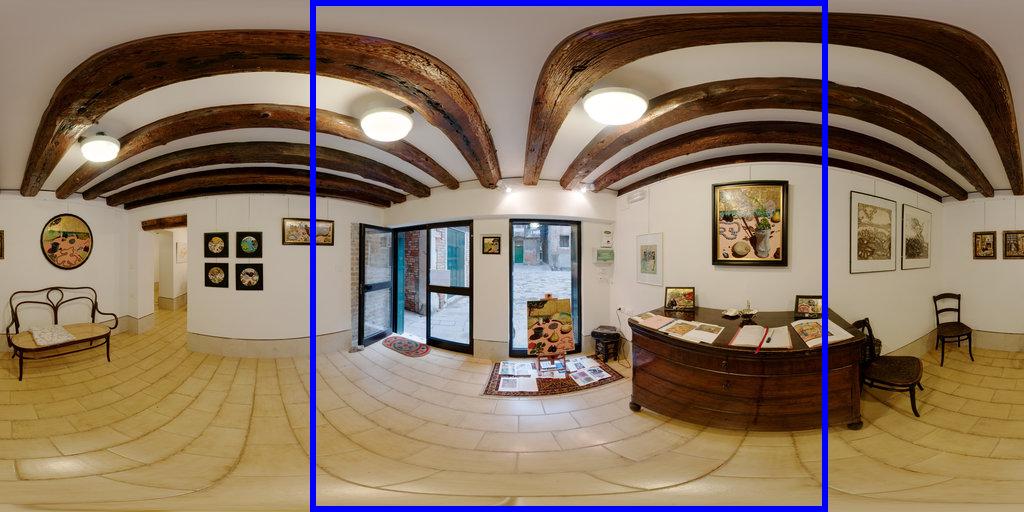}&
\includegraphics[width=0.3\textwidth]{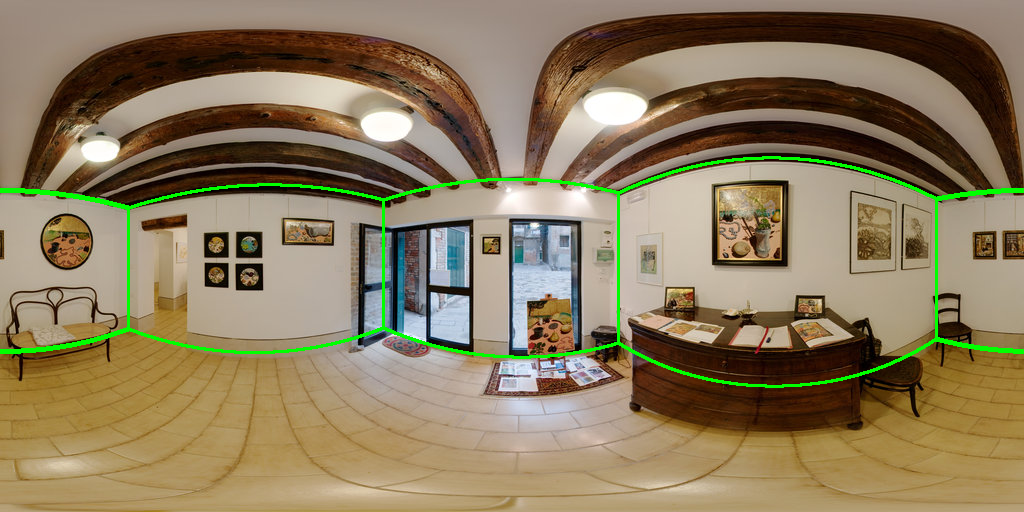}\\
\rotatebox{90}{\hspace{0.15cm} Augmented} & \rotatebox{90}{\hspace{0.45cm} image} &
\includegraphics[width=0.3\textwidth]{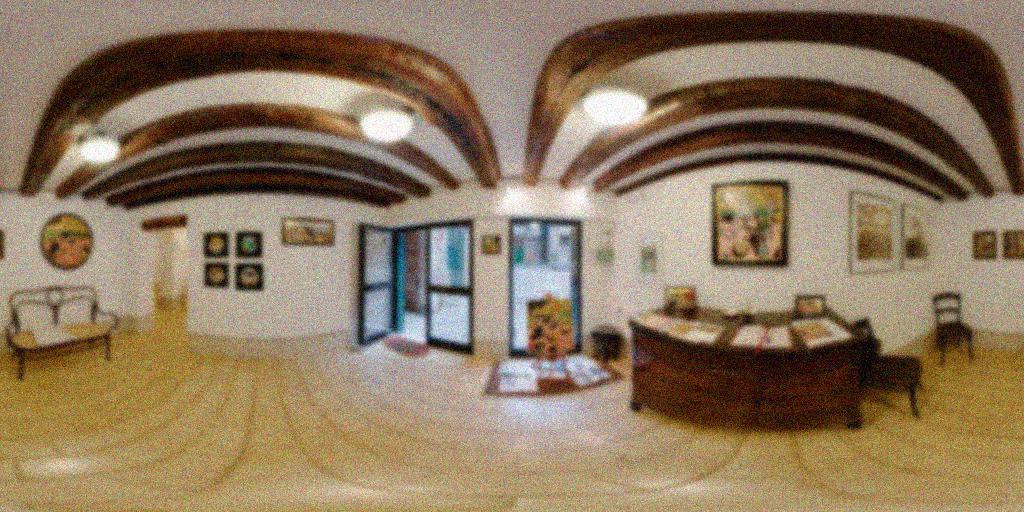}&
 \includegraphics[width=0.3\textwidth]{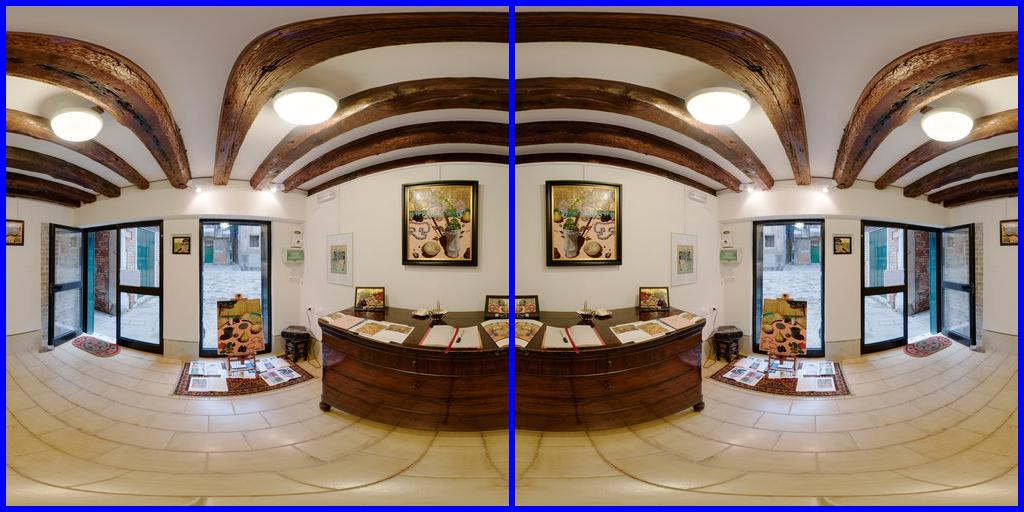}&
 \includegraphics[width=0.3\textwidth]{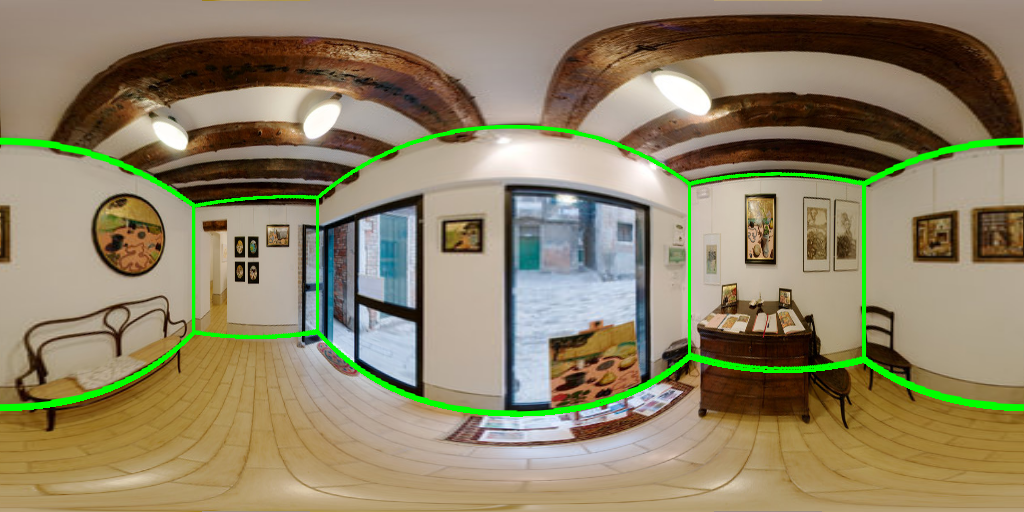}\\ 
%\rotatebox{90}{ground-truth}
&\rotatebox{90}{\hspace{0.4cm} Labels}  &
\includegraphics[width=0.3\textwidth]{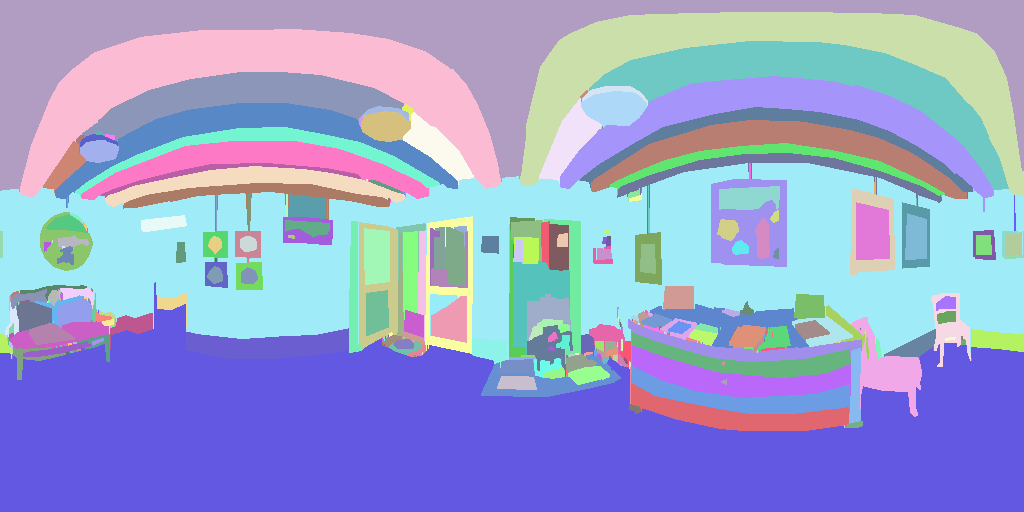}&
\includegraphics[width=0.3\textwidth]{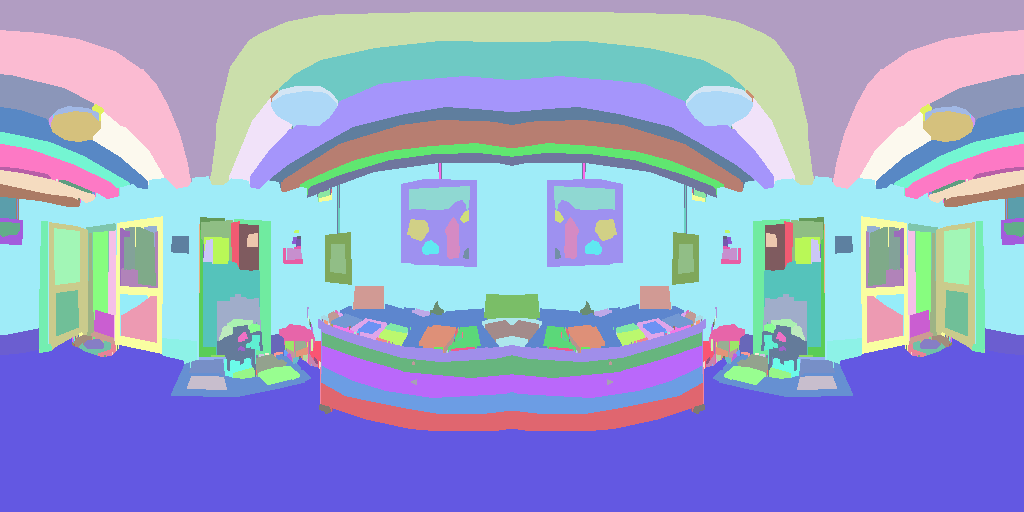}&
 \includegraphics[width=0.3\textwidth]{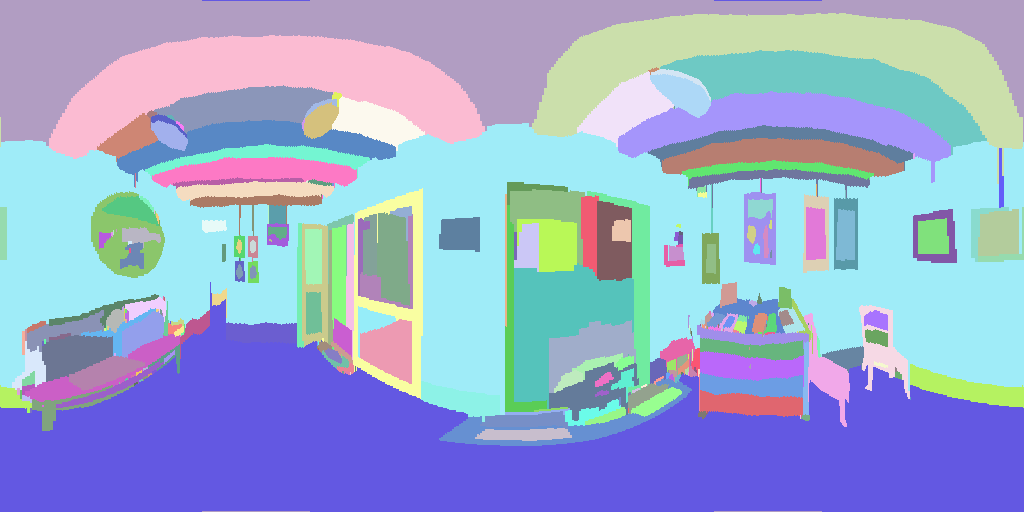}\\[-1ex]
\end{tabular}
}
\caption{Example of data augmentation used during training. 
\textbf{Left:} Standard Gaussian blur and noise (here with respective maximal variance $\sigma=20$ and $\sigma=2$). The ground-truth labels are not impacted. \textbf{Middle:} Crop \& mirror strategy. A random crop of half-width is selected (represented by the green square) and mirrored to form a new \omni image. This method combines horizontal rolling, flipping and also creates information at the mirror border. \textbf{Right:} Panoramic stretch \cite{sun2019horizonnet} to introduce distortions in the \omni image (here with parameters $k_x=0.5$, $k_y=1.25$ that correspond to a respective enlargement and a shrinking of the areas where $|x|\approx1$ and $|y|\approx1$). The layout of the scene is represented by the green lines to more easily apprehend the distortion.} \vspace{-0.1cm}
\label{fig:da}
\end{figure}

To overcome these challenges, we propose to use data augmentation techniques 
tailored explicitly for \omni images.
A straightforward augmentation technique would consist in horizontally rolling the \omni image and its ground-truth \cite{lo2002multiple}.
As stated in~\cite{schubert2019circular}, such data augmentation 
strategy does not bring any diversity in a pure CNN network.
Nevertheless, in our context, 
since average superpixel features are extracted according to an spherical initialization label map, 
a roll of the image may have a different impact on the produced segmentation.
To go further, we also propose to combine random half-width cropping and horizontal mirroring of the input image and ground-truth (see \fig\ref{fig:da}(middle)).
This way, in a single transformation, we combine rolling and flipping while creating information at the mirror border.

Finally, we use the \textit{panoramic stretch} approach of~\cite{sun2019horizonnet} to introduce spatial distortions.
To stretch a \omni image, $[x,y,z]$ coordinates are simply multiplied by a respective factor $[k_x,k_y,k_z]$ and projected back to the sphere.
Pixel values are then computed using bilinear interpolation.
Since setting $k_z$ would affect the projection of $x$ and $y$ values the same way, authors propose to only set $k_x$ and $k_y$ parameters. 
The 3D coordinates maps in \fig\ref{fig:3D_coordinates} represent the image area that would be affected by increasing one of the parameters.
For instance, setting $k_y<1$ would zoom on the region where $y$ values are close to -1 and 1 (see \fig\ref{fig:da}(right)).
We refer the reader to~\cite{sun2019horizonnet} to more details on the stretching algorithm
and to our supp. mat. for additional examples.

With such data augmentation, we are able to greatly enrich the training dataset while preserving the spherical geometry of \omni images.
We demonstrate the improvement of performance obtained using these techniques during training in Section~\ref{subsec:ablation}.

\section{Results}

\subsection{\label{subsec:validation}Validation Framework}

\subsubsection{Datasets}

In our experiments, 
we considered two relevant spherical segmentation datasets
containing various accurately segmented objects
(see examples in supp. mat.).
The first dataset called Panorama Segmentation Dataset (PSD)~\cite{wan2018} 
is the reference one and contains 75 images of $512{\times}1024$ pixels
from the SUN360 dataset~\cite{xiao2012recognizing}.
The ground-truth manual segmentations from~\cite{wan2018}, 
contain an average number of 510 objects 
with an average size of 1334 pixels.
To fairly compare deep learning methods, 
we respectively consider 55, 5 and 15 images for the train, validation and test sets.
In Section \ref{subsec:results}, we also compare the performance
on PSD images affected by an additive white Gaussian noise of variance 20.

To further demonstrate the performance of DSS, 
we choose to consider for the first time in spherical superpixel methods evaluation,
the Wild PAnoramic Semantic Segmentation (WildPASS) dataset \cite{yang2021capturing},
containing 500 omnidirectional natural road images.
We resize the images to $512{\times}1024$ and split the dataset into 
respectively 300, 100 and 100 images for train, validation and test sets.

\vspace{-0.15cm}

\subsubsection{Parameter settings}

Our data augmentation is applied on-the-fly during training.
It includes (i) applying a random Gaussian blur with a 
variance $\sigma\in[0, 2]$, 
(ii) adding Gaussian noise of variance $\sigma\in[0,20]$, 
(iii) random flipping, horizontal rolling and half-width random crop and mirror with a 0.5 probability, 
and (iv) panoramic stretching with random parameters $k_x$ and $k_y$ between 0.5 and 2.
During training, $\lambda=1$ in \eqref{loss} 
and images are downsized to $256{\times}512$ pixels, 
so our model can understand the whole scene's geometry,
contrary to the $201{\times}201$ crops used in~\cite{jampani2018superpixel}.
We refer the reader to the supp. mat. for training details.

\vspace{-0.15cm}

\subsubsection{Evaluation metrics}

The main challenge in superpixel segmentation 
is the ability to produce superpixels that are contained 
into the image objects, with respect to a ground-truth segmentation.
Regularity is also an important aspect to interactive applications or to later extract significant neighborhoods~\cite{giraud2017_jei}.
Since these criteria are generally contradictory, 
efficiently maximizing both is usually the bottleneck of superpixel methods. 
These aspects can be relevantly evaluated with state-of-the-art dedicated metrics
\cite{giraud2017_jei}.
In the following, we denote superpixel segmentation as $\SSS=\{S_i\}$ and ground-truth segmentation as $\GG=\{G_j\}$ with their respective borders $\mathcal{B}(\SSS)$ and $\mathcal{B(\GG)}$. %segmentations as:

The mainly reported measure is the segmentation accuracy, with the Achievable Segmentation Accuracy (ASA)~\cite{liu2011} such that: \vspace{-0.05cm}

{\small
\begin{equation}
 \text{ASA}(\SSS,\GG) = \frac{1}{\sum\limits_{S_i \in \SSS}|S_i|}\sum_{S_i}\underset{G_j\in \GG}{\max}|S_i\cap G_j|.  \label{asa}
\end{equation}}

%\vspace{-0.05cm}
This aspect can also be evaluated by focusing on the contour adherence of superpixels to the object borders, using the Boundary-Recall (BR) such that: \vspace{-0.05cm}

{\small
\begin{equation}
\text{BR}(\SSS,\GG) = \frac{1}{|\mathcal{B}(\GG)|}\sum_{p\in\mathcal{B}(\GG)}\delta[\min_{q\in\mathcal{B}(\SSS)}\|p-q\|< \epsilon]  ,   \label{br}
\end{equation}}%  \vspace{-0.1cm}
\noindent with 
 $\epsilon$ a distance threshold set to $2$ pixels \cite{giraud2017_jei}, 
 and $\delta[a]=1$ when $a$ is true and $0$ otherwise.
Since it only measures recall, BR should be compared to the Contour Density (CD), \emph{i.e.}, the proportion of border pixels of the generated superpixels.

Finally, to evaluate the regularity aspect, we use
the Generalized Global Regularity (GGR) metric 
that adapts the metric proposed in~\cite{giraud2017_jei}
to \omni images~\cite{giraud2023sphsps}. 
This metric evaluates the convexity, balanced pixel distribution, contour smoothness of each shape and also how homogeneous the shape distribution is within the segmentation.
We refer the reader to~\cite{giraud2023sphsps} to more details on the GGR metric.

\subsection{\label{subsec:ablation}Ablation Study}

In Table~\ref{table:ablation}, we report the impact of each data augmentation strategy and the spherical CNN architecture, \emph{i.e.}, using circular padding instead of zero padding~\cite{jampani2018superpixel} on the PSD and noisy PSD images for an average number of $K=500$ superpixels.
Each augmentation strategy increases the training efficiency in terms of segmentation accuracy, while the circular padding logically improves the spherical regularity
by cancelling the artificial horizontal border of the segmentation.
This confirms the interest of improving the original SSN method with spherical CNN architecture and specific augmentation strategies for 360° images.

\begin{table}[t]
\centering
\newcommand{\anglee}{25}
\newcommand{\spacew}{2mm}
\caption{{Ablation study of the proposed DSS method on PSD and noisy PSD images on ASA ($\uparrow$), CD/BR ($\downarrow$) and GGR ($\uparrow$).  CD is given for BR=0.8. Best and second best results are respectively in bold and underlined font.}} \vspace{-0.1cm}
{\scriptsize
  \begin{tabular}{@{\hspace{0mm}}c@{\hspace{\spacew}}c@{\hspace{\spacew}}c@{\hspace{0mm}}cc@{\hspace{2mm}}c@{\hspace{4mm}}c@{\hspace{\spacew}}c@{\hspace{\spacew}}c@{\hspace{4mm}}c@{\hspace{\spacew}}c@{\hspace{\spacew}}c@{\hspace{0mm}}}
  \multicolumn{3}{c@{\hspace{0mm}}}{\color{gray} Data augmentation}\\[0.5ex] \cline{1-3} \\[-2ex]
    {Gaussian}&     {Horizontal}&      {Panoramic}& &&    {Circular} &
    \multicolumn{3}{c}{PSD} & \multicolumn{3}{c}{Noisy PSD}\\
    \cline{7-12}
  {blur\&noise}&
   {crop\&mirror}&
    {strecth}&&&
     {padding}&
       ASA & CD/BR & GGR & ASA & CD/BR & GGR\\
       \hline
- & - & - &&& \checkmark & 0.862 & 0.134 & 0.385 & 0.858 & 0.139 &0.386\\
\checkmark & - & - &&& \checkmark & 0.877 & \underline{0.119}  & \textbf{0.444} & 0.868 & 0.132 & \textbf{0.461}\\
\checkmark & \checkmark & - &&& \checkmark & \underline{0.888} &  \textbf{0.117} & \underline{0.413} & 0.883 &  \textbf{0.124}  & \underline{0.423} \\
\checkmark &\checkmark & \checkmark &&& - & 0.887 & {0.124} & {0.387} & \underline{0.884} & {0.134} & {0.390} \\
\checkmark & \checkmark & \checkmark &&& \checkmark &  \textbf{0.890} &  {0.122} &  {0.388} & \textbf{0.886}&  \underline{0.132} & {0.392} \\ \hline
  \end{tabular}}%
  \label{table:ablation}
\end{table}

\subsection{Comparison to State-of-the-Art Methods\label{subsec:results}}

\subsubsection{Compared methods}
In our experiments,
we compare DSS
to the spherical methods:
SSLIC \cite{zhao2018},
SSNIC \cite{da2021fast},
SphLSC 
and
SphSPS \cite{giraud2023sphsps}.
We also compare to some recent planar methods:
LSC \cite{chen2017},
SNIC \cite{achanta2017superpixels},
GMMSP \cite{Ban18}, 
and
SSN \cite{jampani2018superpixel}
All methods are used with the default regularity parameters.
For the SSLIC method \cite{zhao2018}, that does not have one, we use a 
color weight of 20 to try to optimize its segmentation accuracy.
For SSN \cite{jampani2018superpixel}, 
we compare to both the initial network trained on the BSD dataset \cite{martin2001} containing planar natural images (SSN-BSD) and to a retrained network on the targeted dataset (SSN-PSD, SSN-WP).

\subsubsection{Evaluation of performance}
\begin{figure}[t]
\centering
\newcommand{\hhh}{0.22\textwidth}
{\scriptsize 
\begin{tabular}{@{\hspace{0mm}}c@{\hspace{1mm}}c@{\hspace{3mm}}c@{\hspace{1mm}}c@{\hspace{0mm}}}
\includegraphics[height=\hhh,width=0.32\textwidth]{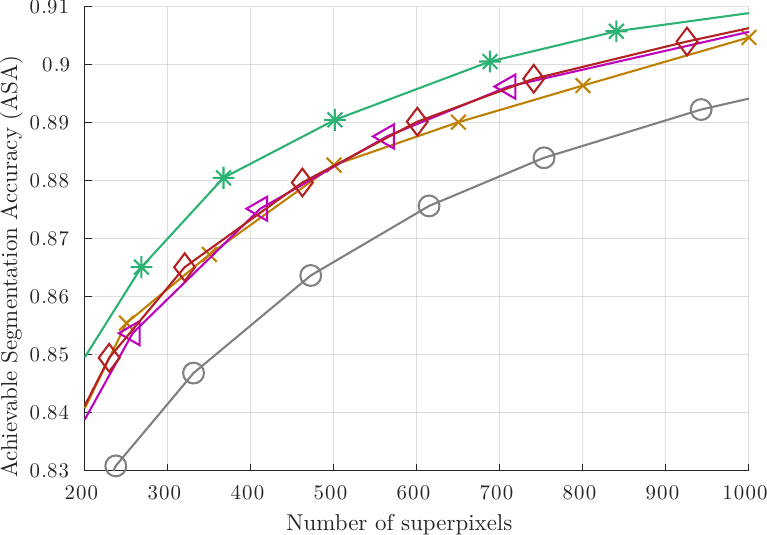} &
\includegraphics[height=\hhh,width=0.32\textwidth]{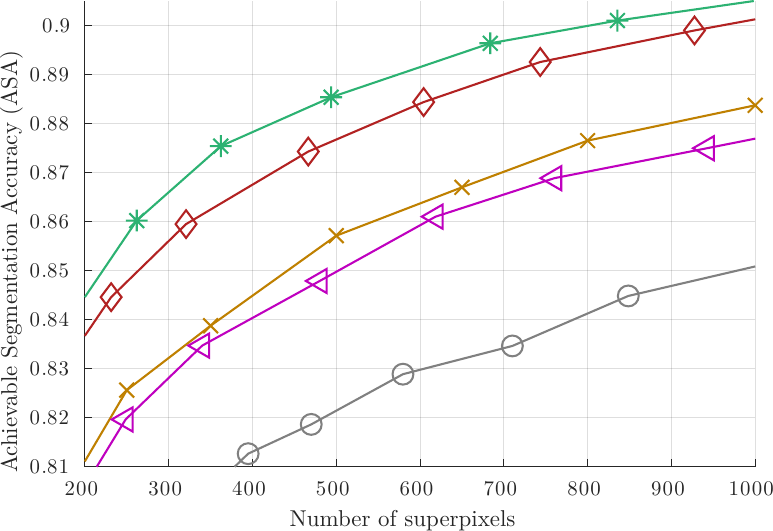} & 
\includegraphics[height=\hhh,width=0.32\textwidth]{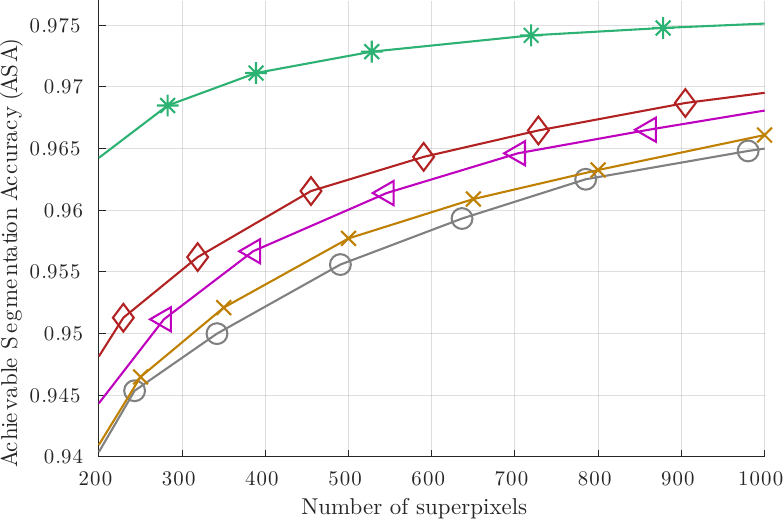}\\
\includegraphics[height=\hhh,width=0.32\textwidth]{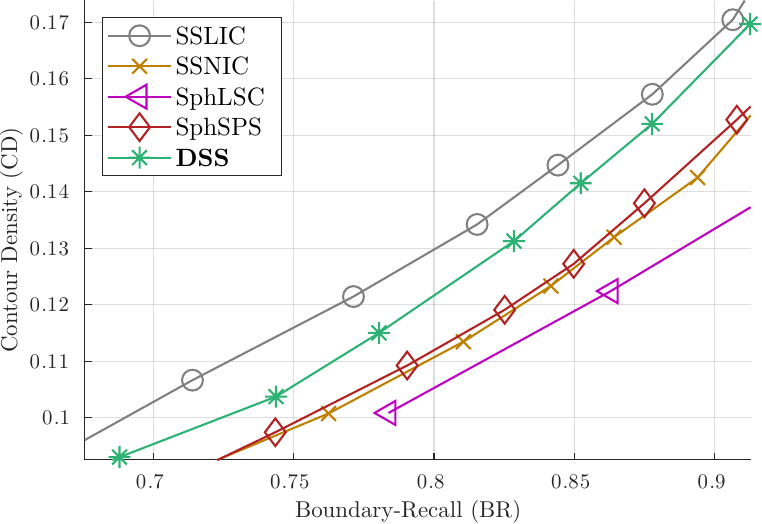} &
\includegraphics[height=\hhh,width=0.32\textwidth]{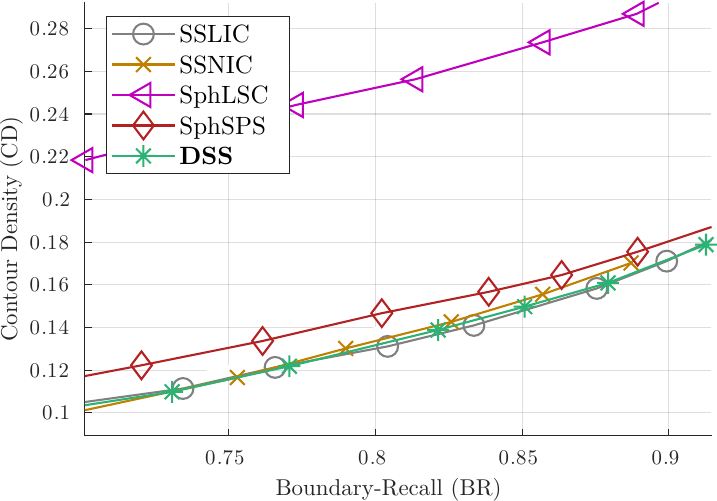} & 
\includegraphics[height=\hhh,width=0.32\textwidth]{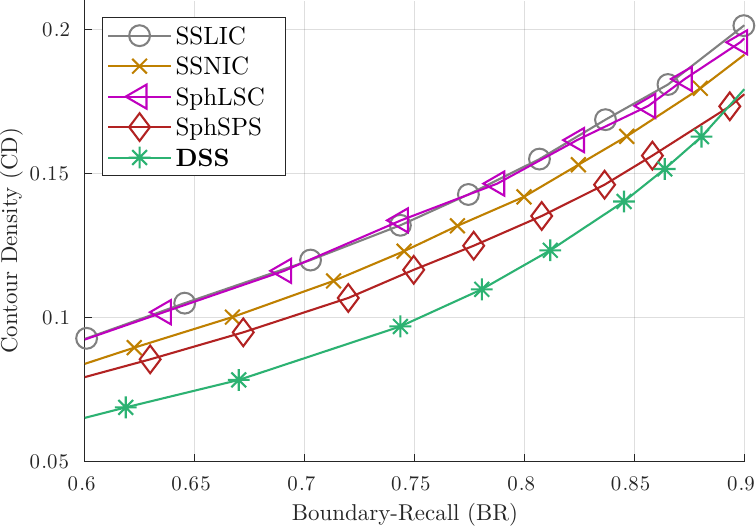}\\ 
(a) PSD images & (b) Noisy PSD images & (c) WP images  \\[-0.5ex]
\end{tabular}}%
 \caption{Comparison of DSS to state-of-the-art methods. \textbf{Top:} Segmentation accuracy evaluated with ASA \eqref{asa}. \textbf{Bottom:} 
 Contour adherence in terms of CD vs BR \eqref{br}. \vspace{-0.1cm}
 }
 \label{fig:results_soa}
\end{figure}

We compare the proposed DSS to the spherical methods in terms of segmentation accuracy (ASA) and also contour adherence (CD/BR) for several superpixel numbers required $K$, 
on the PSD images \fig\ref{fig:results_soa}(a), 
noisy PSD images \fig\ref{fig:results_soa}(b) 
and WP images \fig\ref{fig:results_soa}(c).

We observe that DSS obtains the highest segmentation accuracy (ASA) on all type of images.
We can also see that our method is robust to noise contrary to most state-of-the-art methods that present a significant loss of performance on such slightly altered images.
Finally, we can note that DSS superpixels also have the highest contour adherence (lowest CD/BR) compared to other methods, only except on noise-free PSD images.
This can be simply explained by the fact that our method, as SSN, 
does not explicitly integrate a contour adherence loss
and that the ground-truth segmentations in the PSD dataset contain
many annotations of very thin objects that impact such metric.

In Table~\ref{table:soa}, we report results for $K=500$, 
also including the regularity metric (GGR), and performance obtained with planar methods.
We observe that GGR discriminates well the planar and spherical methods.
DSS is among the spherical methods, 
having higher spherical regularity than planar methods, 
and it also preserves its regularity in the presence of noise.

Compared to SSN,
we can first notice that SSN trained on the BSD
does not generalize very well when applied on PSD or WP images.
It demonstrates the capacity of CNNs to extract semantic information
and that performance of generalization
may highly depend on the similarity of annotations.
We also observe that DSS slightly outperforms
SSN retrained on the PSD and WP datasets,
in terms of segmentation accuracy.
SSN is able to train its network by providing image crops, which is
a much more efficient learning strategy than to provide the whole image, 
as we have to do in DSS.
Nevertheless, with our data augmentation strategy, 
we can maintain the same level of accuracy while 
generating spherical superpixels that may follow the deformed objects.

\begin{figure*}[ht!]
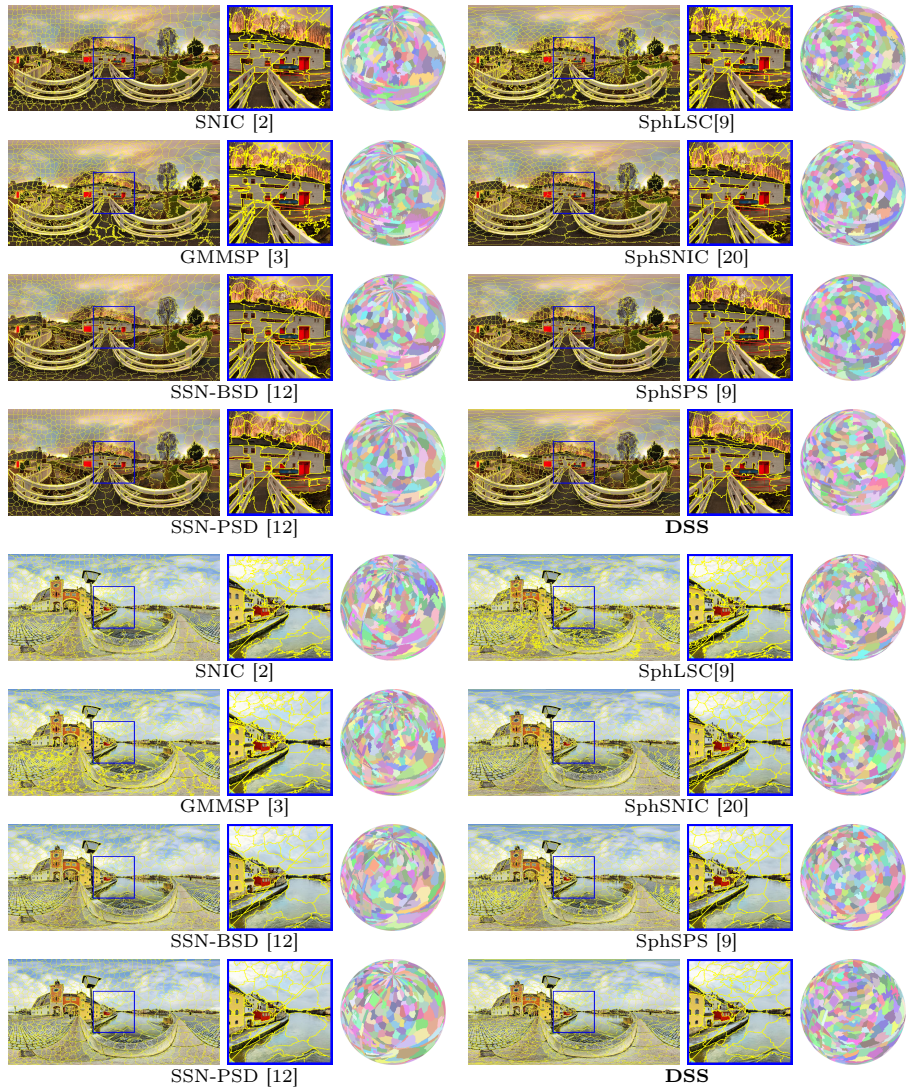

\centering
{\scriptsize 
\newcommand{\ww}{0.23\textwidth} 
\newcommand{\pppr}{0.115\textwidth} 
\newcommand{\ppp}{0.115\textwidth}   
\begin{tabular}{@{\hspace{1mm}}c@{\hspace{1mm}}c@{\hspace{1mm}}c@{\hspace{3mm}}c@{\hspace{1mm}}c@{\hspace{1mm}}c@{\hspace{0mm}}}
\includegraphics[width=\ww,height=\pppr]{res_snic_\thenumb}&
\includegraphics[width=\ppp,height=\pppr]{res_snic_\thenumb_zoom}&
\includegraphics[width=\ppp,height=\pppr]{res_snic_\thenumb_sphere}&
\includegraphics[width=\ww,height=\pppr]{res_sphlsc_\thenumb}&
\includegraphics[width=\ppp,height=\pppr]{res_sphlsc_\thenumb_zoom}&
\includegraphics[width=\ppp,height=\pppr]{res_sphlsc_\thenumb_sphere}
\\[-0.5ex]
\multicolumn{3}{c}{SNIC \cite{achanta2017superpixels}}& 
\multicolumn{3}{c}{{SphLSC\cite{giraud2023sphsps}}} \\[0.75ex]
\includegraphics[width=\ww,height=\pppr]{res_gmmsp_\thenumb}&
\includegraphics[width=\ppp,height=\pppr]{res_gmmsp_\thenumb_zoom}&
\includegraphics[width=\ppp,height=\pppr]{res_gmmsp_\thenumb_sphere}&
\includegraphics[width=\ww,height=\pppr]{res_sphsnic_\thenumb}&
\includegraphics[width=\ppp,height=\pppr]{res_sphsnic_\thenumb_zoom}&
\includegraphics[width=\ppp,height=\pppr]{res_sphsnic_\thenumb_sphere}
\\[-0.5ex]
\multicolumn{3}{c}{GMMSP \cite{Ban18}}& 
\multicolumn{3}{c}{{SphSNIC \cite{da2021fast}}} \\[0.75ex]
\includegraphics[width=\ww,height=\pppr]{res_ssn_\thenumb}&
\includegraphics[width=\ppp,height=\pppr]{res_ssn_\thenumb_zoom}&
\includegraphics[width=\ppp,height=\pppr]{res_ssn_\thenumb_sphere}&
\includegraphics[width=\ww,height=\pppr]{res_glips_\thenumb}&
\includegraphics[width=\ppp,height=\pppr]{res_glips_\thenumb_zoom}&
\includegraphics[width=\ppp,height=\pppr]{res_glips_\thenumb_sphere}
\\[-0.5ex]
\multicolumn{3}{c}{SSN-BSD \cite{jampani2018superpixel}}&
\multicolumn{3}{c}{{{SphSPS \cite{giraud2023sphsps}}}} \\[0.75ex]
\includegraphics[width=\ww,height=\pppr]{res_ssn_re_\thenumb}&
\includegraphics[width=\ppp,height=\pppr]{res_ssn_re_\thenumb_zoom}&
\includegraphics[width=\ppp,height=\pppr]{res_ssn_re_\thenumb_sphere}&
\includegraphics[width=\ww,height=\pppr]{res_dss_\thenumb}&
\includegraphics[width=\ppp,height=\pppr]{res_dss_\thenumb_zoom}&
\includegraphics[width=\ppp,height=\pppr]{res_dss_\thenumb_sphere}
\\[-0.5ex]
\multicolumn{3}{c}{SSN-PSD \cite{jampani2018superpixel}}&
\multicolumn{3}{c}{\textbf{DSS}} 
\setcounter{numb}{63} \\[2ex] 
%
% Exemple 2
%
\includegraphics[width=\ww,height=\pppr]{res_snic_\thenumb}&
\includegraphics[width=\ppp,height=\pppr]{res_snic_\thenumb_zoom}&
\includegraphics[width=\ppp,height=\pppr]{res_snic_\thenumb_sphere}&
\includegraphics[width=\ww,height=\pppr]{res_sphlsc_\thenumb}&
\includegraphics[width=\ppp,height=\pppr]{res_sphlsc_\thenumb_zoom}&
\includegraphics[width=\ppp,height=\pppr]{res_sphlsc_\thenumb_sphere}
\\[-0.5ex]
\multicolumn{3}{c}{SNIC \cite{achanta2017superpixels}}& 
\multicolumn{3}{c}{{SphLSC\cite{giraud2023sphsps}}} \\[0.75ex]
\includegraphics[width=\ww,height=\pppr]{res_gmmsp_\thenumb}&
\includegraphics[width=\ppp,height=\pppr]{res_gmmsp_\thenumb_zoom}&
\includegraphics[width=\ppp,height=\pppr]{res_gmmsp_\thenumb_sphere}&
\includegraphics[width=\ww,height=\pppr]{res_sphsnic_\thenumb}&
\includegraphics[width=\ppp,height=\pppr]{res_sphsnic_\thenumb_zoom}&
\includegraphics[width=\ppp,height=\pppr]{res_sphsnic_\thenumb_sphere}
\\[-0.5ex]
\multicolumn{3}{c}{GMMSP \cite{Ban18}}& 
\multicolumn{3}{c}{{SphSNIC \cite{da2021fast}}} \\[0.75ex]
\includegraphics[width=\ww,height=\pppr]{res_ssn_\thenumb}&
\includegraphics[width=\ppp,height=\pppr]{res_ssn_\thenumb_zoom}&
\includegraphics[width=\ppp,height=\pppr]{res_ssn_\thenumb_sphere}&
\includegraphics[width=\ww,height=\pppr]{res_glips_\thenumb}&
\includegraphics[width=\ppp,height=\pppr]{res_glips_\thenumb_zoom}&
\includegraphics[width=\ppp,height=\pppr]{res_glips_\thenumb_sphere}
\\[-0.5ex]
\multicolumn{3}{c}{SSN-BSD \cite{jampani2018superpixel}}&
\multicolumn{3}{c}{{{SphSPS \cite{giraud2023sphsps}}}} \\[0.75ex]
\includegraphics[width=\ww,height=\pppr]{res_ssn_re_\thenumb}&
\includegraphics[width=\ppp,height=\pppr]{res_ssn_re_\thenumb_zoom}&
\includegraphics[width=\ppp,height=\pppr]{res_ssn_re_\thenumb_sphere}&
\includegraphics[width=\ww,height=\pppr]{res_dss_\thenumb}&
\includegraphics[width=\ppp,height=\pppr]{res_dss_\thenumb_zoom}&
\includegraphics[width=\ppp,height=\pppr]{res_dss_\thenumb_sphere}
\\[-0.5ex]
\multicolumn{3}{c}{SSN-PSD \cite{jampani2018superpixel}}&
\multicolumn{3}{c}{\textbf{DSS}} \\[-1ex]
\end{tabular}}
\caption{Qualitative comparison on PSD images, for planar (left) and spherical methods (right)
for two superpixel numbers $K=1200$ (top-left) and $K=400$ (bottom right).
 }  \vspace{-0.2cm}
\label{fig:psd}
\end{figure*}

\begin{figure*}[ht!]
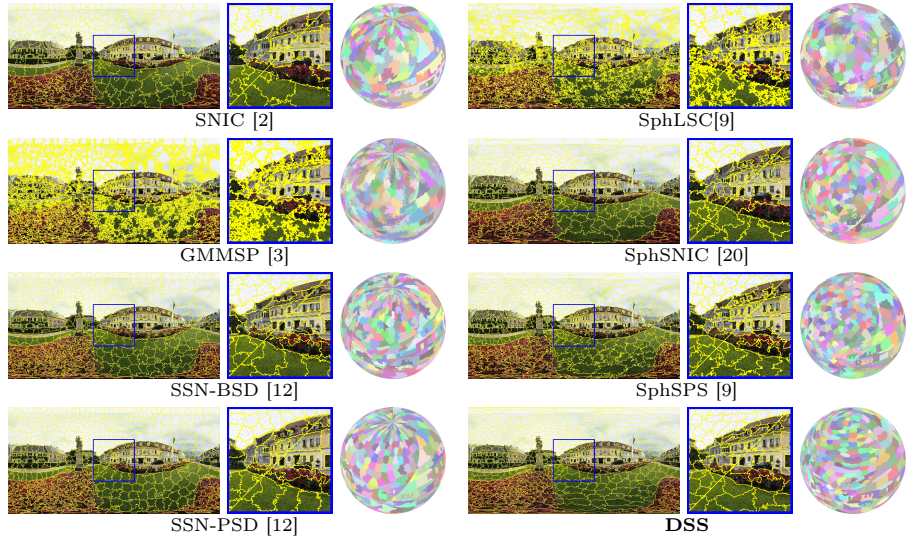

\centering
{\scriptsize 
\newcommand{\ww}{0.23\textwidth} 
\newcommand{\pppr}{0.115\textwidth} 
\newcommand{\ppp}{0.115\textwidth}   
\begin{tabular}{@{\hspace{1mm}}c@{\hspace{1mm}}c@{\hspace{1mm}}c@{\hspace{3mm}}c@{\hspace{1mm}}c@{\hspace{1mm}}c@{\hspace{0mm}}}
\includegraphics[width=\ww,height=\pppr]{res_snic_\thenumb_noise}&
\includegraphics[width=\ppp,height=\pppr]{res_snic_\thenumb_noise_zoom}&
\includegraphics[width=\ppp,height=\pppr]{res_snic_\thenumb_noise_sphere}&
\includegraphics[width=\ww,height=\pppr]{res_sphlsc_\thenumb_noise}&
\includegraphics[width=\ppp,height=\pppr]{res_sphlsc_\thenumb_noise_zoom}&
\includegraphics[width=\ppp,height=\pppr]{res_sphlsc_\thenumb_noise_sphere}
\\[-0.5ex]
\multicolumn{3}{c}{SNIC \cite{achanta2017superpixels}}& 
\multicolumn{3}{c}{{SphLSC\cite{giraud2023sphsps}}} \\[0.75ex]
\includegraphics[width=\ww,height=\pppr]{res_gmmsp_\thenumb_noise}&
\includegraphics[width=\ppp,height=\pppr]{res_gmmsp_\thenumb_noise_zoom}&
\includegraphics[width=\ppp,height=\pppr]{res_gmmsp_\thenumb_noise_sphere}&
\includegraphics[width=\ww,height=\pppr]{res_sphsnic_\thenumb_noise}&
\includegraphics[width=\ppp,height=\pppr]{res_sphsnic_\thenumb_noise_zoom}&
\includegraphics[width=\ppp,height=\pppr]{res_sphsnic_\thenumb_noise_sphere}
\\[-0.5ex]
\multicolumn{3}{c}{GMMSP \cite{Ban18}}& 
\multicolumn{3}{c}{{SphSNIC \cite{da2021fast}}} \\[0.75ex]
\includegraphics[width=\ww,height=\pppr]{res_ssn_\thenumb_noise}&
\includegraphics[width=\ppp,height=\pppr]{res_ssn_\thenumb_noise_zoom}&
\includegraphics[width=\ppp,height=\pppr]{res_ssn_\thenumb_noise_sphere}&
\includegraphics[width=\ww,height=\pppr]{res_glips_\thenumb_noise}&
\includegraphics[width=\ppp,height=\pppr]{res_glips_\thenumb_noise_zoom}&
\includegraphics[width=\ppp,height=\pppr]{res_glips_\thenumb_noise_sphere}
\\[-0.5ex]
\multicolumn{3}{c}{SSN-BSD \cite{jampani2018superpixel}}&
\multicolumn{3}{c}{{{SphSPS \cite{giraud2023sphsps}}}} \\[0.75ex]
\includegraphics[width=\ww,height=\pppr]{res_ssn_re_\thenumb_noise}&
\includegraphics[width=\ppp,height=\pppr]{res_ssn_re_\thenumb_noise_zoom}&
\includegraphics[width=\ppp,height=\pppr]{res_ssn_re_\thenumb_noise_sphere}&
\includegraphics[width=\ww,height=\pppr]{res_dss_\thenumb_noise}&
\includegraphics[width=\ppp,height=\pppr]{res_dss_\thenumb_noise_zoom}&
\includegraphics[width=\ppp,height=\pppr]{res_dss_\thenumb_noise_sphere}
\\[-0.5ex]
\multicolumn{3}{c}{SSN-PSD \cite{jampani2018superpixel}}&
\multicolumn{3}{c}{\textbf{DSS}} \\[-1ex]
\end{tabular}}
\caption{Qualitative comparison on a noisy PSD image for planar (left) and spherical methods (right)
for two superpixel numbers $K=1200$ (top-left) and $K=400$ (bottom right).
DSS is able to preserve its regularity and accuracy compared to most methods.
 }  \vspace{-0.2cm}
\label{fig:noisy_psd}
\end{figure*}

\begin{figure*}[ht!]
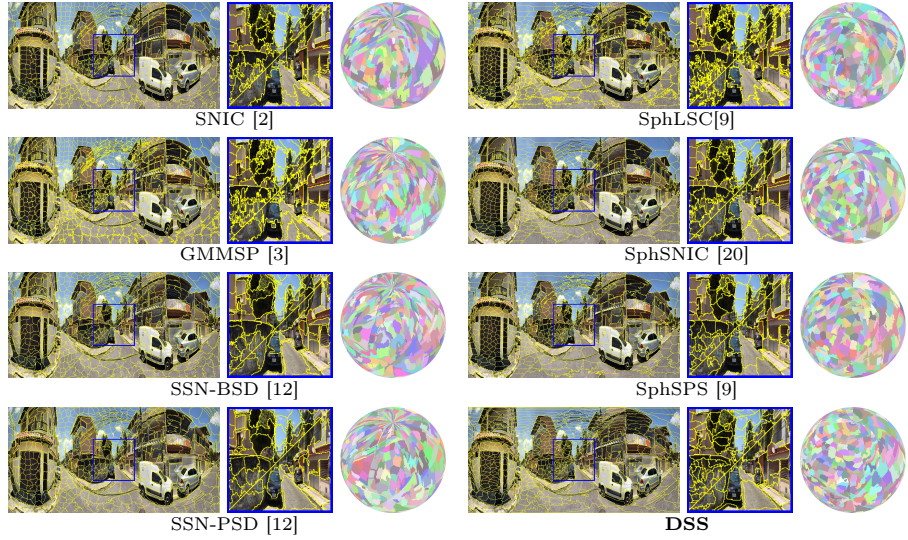

\centering
{\scriptsize 
\newcommand{\ww}{0.23\textwidth} 
\newcommand{\pppr}{0.115\textwidth} 
\newcommand{\ppp}{0.115\textwidth}   
\begin{tabular}{@{\hspace{1mm}}c@{\hspace{1mm}}c@{\hspace{1mm}}c@{\hspace{3mm}}c@{\hspace{1mm}}c@{\hspace{1mm}}c@{\hspace{0mm}}}
\includegraphics[width=\ww,height=\pppr]{res_snic_\thenumb}&
\includegraphics[width=\ppp,height=\pppr]{res_snic_\thenumb_zoom}&
\includegraphics[width=\ppp,height=\pppr]{res_snic_\thenumb_sphere}&
\includegraphics[width=\ww,height=\pppr]{res_sphlsc_\thenumb}&
\includegraphics[width=\ppp,height=\pppr]{res_sphlsc_\thenumb_zoom}&
\includegraphics[width=\ppp,height=\pppr]{res_sphlsc_\thenumb_sphere}
\\[-0.5ex]
\multicolumn{3}{c}{SNIC \cite{achanta2017superpixels}}& 
\multicolumn{3}{c}{{SphLSC\cite{giraud2023sphsps}}} \\[0.75ex]
\includegraphics[width=\ww,height=\pppr]{res_gmmsp_\thenumb}&
\includegraphics[width=\ppp,height=\pppr]{res_gmmsp_\thenumb_zoom}&
\includegraphics[width=\ppp,height=\pppr]{res_gmmsp_\thenumb_sphere}&
\includegraphics[width=\ww,height=\pppr]{res_sphsnic_\thenumb}&
\includegraphics[width=\ppp,height=\pppr]{res_sphsnic_\thenumb_zoom}&
\includegraphics[width=\ppp,height=\pppr]{res_sphsnic_\thenumb_sphere}
\\[-0.5ex]
\multicolumn{3}{c}{GMMSP \cite{Ban18}}& 
\multicolumn{3}{c}{{SphSNIC \cite{da2021fast}}} \\[0.75ex]
\includegraphics[width=\ww,height=\pppr]{res_ssn_\thenumb}&
\includegraphics[width=\ppp,height=\pppr]{res_ssn_\thenumb_zoom}&
\includegraphics[width=\ppp,height=\pppr]{res_ssn_\thenumb_sphere}&
\includegraphics[width=\ww,height=\pppr]{res_glips_\thenumb}&
\includegraphics[width=\ppp,height=\pppr]{res_glips_\thenumb_zoom}&
\includegraphics[width=\ppp,height=\pppr]{res_glips_\thenumb_sphere}
\\[-0.5ex]
\multicolumn{3}{c}{SSN-BSD \cite{jampani2018superpixel}}&
\multicolumn{3}{c}{{{SphSPS \cite{giraud2023sphsps}}}} \\[0.75ex]
\includegraphics[width=\ww,height=\pppr]{res_ssn_re_\thenumb}&
\includegraphics[width=\ppp,height=\pppr]{res_ssn_re_\thenumb_zoom}&
\includegraphics[width=\ppp,height=\pppr]{res_ssn_re_\thenumb_sphere}&
\includegraphics[width=\ww,height=\pppr]{res_dss_\thenumb}&
\includegraphics[width=\ppp,height=\pppr]{res_dss_\thenumb_zoom}&
\includegraphics[width=\ppp,height=\pppr]{res_dss_\thenumb_sphere}
\\[-0.5ex]
\multicolumn{3}{c}{SSN-PSD \cite{jampani2018superpixel}}&
\multicolumn{3}{c}{\textbf{DSS}} \\[-1ex]
\end{tabular}}
\caption{Qualitative comparison on a WP image for planar (left) and spherical methods (right)
for two superpixel numbers $K=1200$ (top-left) and $K=400$ (bottom right). Note how DSS is able to capture the car in the image center.
 }  \vspace{-0.2cm}
\label{fig:wp}
\end{figure*}

Finally, qualitative results are respectively shown on PSD, noisy PSD and WP images in \fig\ref{fig:psd},~\ref{fig:noisy_psd},~\ref{fig:wp}.
For planar methods, we can note the projection irregularity around the sphere's poles.
DSS produces spherically regular superpixels that well capture the image objects.

\begin{table}[t]
\caption{Quantitative comparison of DSS to state-of-the-art methods for an average number of $K=500$ superpixels on ASA ($\uparrow$), CD/BR ($\downarrow$) and GGR ($\uparrow$). CD is given for BR=0.8. Best and second best results are respectively in bold and underlined font.} \vspace{-0.15cm}
\centering
{\scriptsize
    \begin{tabular}{@{\hspace{0mm}}l@{\hspace{2mm}}l@{\hspace{2mm}}c@{\hspace{2mm}}c@{\hspace{2mm}}c@{\hspace{4mm}}c@{\hspace{2mm}}c@{\hspace{2mm}}c@{\hspace{4mm}}c@{\hspace{2mm}}c@{\hspace{2mm}}c@{\hspace{0mm}}}
   & & \multicolumn{3}{@{\hspace{-4mm}}c@{\hspace{0mm}}}{PSD}
    & \multicolumn{3}{@{\hspace{-6mm}}c@{\hspace{0mm}}}{Noisy PSD}
    & \multicolumn{3}{@{\hspace{0mm}}c@{\hspace{0mm}}}{WP} \\
    \cline{3-11} \\[-2.25ex]
   && ASA & CD/BR & GGR  &ASA & CD/BR & GGR  &ASA & CD/BR & GGR  \\
   \hline
   \\[-2.25ex]
\multirow{6}{*}{\rotatebox{90}{\color{gray} \text{ } Planar}} & 
LSC \cite{chen2017}&   0.877 &0.138&0.347&0.844& 0.303 &0.334 & 0.962&0.153& 0.313\\
&SNIC \cite{achanta2017superpixels}&0.864 &0.129& 0.361& 0.852&0.139& 0.357 & 0.958 & 0.146 & 0.322 \\
%&SCALP \cite{giraud2018_scalp}&0. &XXX&XXX&XXX&XXX&XXX\\
&GMMSP \cite{Ban18}&0.877 &0.136& 0.339& 0.849&0.329&0.328&0.963 & 0.157 & 0.306\\
&SSN-BSD \cite{jampani2018superpixel}&0.879 & 0.119 &0.328&0.863&0.147&0.321 &0.967 & 0.134 & 0.296 \\
&SSN-PSD/WP   \cite{jampani2018superpixel}&\underline{0.887}&0.114&0.334&0.873&0.141&0.328 & \underline{0.972} & \underline{0.120} & 0.303\\
\hline
   \\[-2.25ex]
\multirow{5}{*}{\rotatebox{90}{\color{gray}Spherical}} &SSLIC \cite{zhao2018}&0.866&0.130 &0.421&0.821&\textbf{0.130}&0.383&0.956&0.152&0.399 \\
&SSNIC \cite{da2021fast}&0.883 & \underline{0.110} & \textbf{0.462}& 0.857 & 0.134 & \textbf{0.399} & 0.958 & 0.142 & \underline{0.410}\\ 
&SphLSC \cite{giraud2023sphsps}&0.882 & \textbf{0.105} & 0.397& 0.850& 0.252 & 0.357 & 0.960 & 0.152 & 0.360 \\
&SphSPS \cite{giraud2023sphsps}&0.883&0.112&\underline{0.452}&\underline{0.877}&0.146&{0.389}& 0.962 & 0.133 & \textbf{0.411}\\
&\textbf{DSS} &\textbf{0.890} & 0.122 & 0.388&\textbf{0.886}&\underline{0.132}&\underline{0.392} & \textbf{0.973} & \textbf{0.118} & 0.356 \\
   \hline 
    \end{tabular}
    } \vspace{-0.15cm}
    \label{table:soa}
\end{table}

\section{Conclusion}

In this work, we proposed DSS, 
the first deep learning-based spherical superpixel segmentation method.
The proposed approach leverages on spherical CNN architectures dedicated to omnidirectional images having a circular geometry.  
We demonstrated that combining a deep learning strategy that respects the spherical geometry along with appropriate data augmentation enables to 
achieve higher and more robust segmentation performance than
both traditional and deep learning-based methods. 

We firmly believe that the presented work holds significant value for the community, given the importance of achieving both accurate segmentation and high regularity in the acquisition space, here spherical, 
for an effective display and processing of adjacent relationships
in computer vision preprocessing tasks.

\newcounter{numb}
\setcounter{numb}{62}

\setcounter{numb}{73} 

\setcounter{numb}{402}

\bibliographystyle{splncs04}
\bibliography{biblio}

\end{document}